\pdfoutput=1
\documentclass[11pt,onecolumn]{article}

\usepackage{files/cvpr}
\usepackage{amsmath, amssymb, graphicx, xfrac, amsfonts, amsbsy, algorithm, color, subfig, cite, url, multirow}
\usepackage[noend]{algpseudocode}
\usepackage[section]{placeins}

\cvprfinalcopy

\def \X {\mathbf{X}}
\def \x {\mathbf{x}}
\def \e {\mathbf{e}}
\def \th {\pmb{\theta}}
\def \ph {\pmb{\phi}}

\def \Rspace { \mathbb{R}}

\def \V {\mathcal{V}}
\def \G {\mathcal{G}}
\def \E {\mathcal{E}}

\begin{document}
%
%%%%%%%%%%%%%%%%%%%%%%%%%%%%%%%%%%%%%%%%%%%%%
%% Title
%%%%%%%%%%%%%%%%%%%%%%%%%%%%%%%%%%%%%%%%%%%%%
\title{ Learning Spatial Relationships between Samples of Patent Image Shapes }

%%%%%%%%%%%%%%%%%%%%%%%%%%%%%%%%%%%%%%%%%%%%%
%% Authors
%%%%%%%%%%%%%%%%%%%%%%%%%%%%%%%%%%%%%%%%%%%%%

\author{ Juan Castorena, Manish Bhattarai and Diane Oyen
	\thanks{J. Castorena, M. Bhattarai and D. Oyen are with the Los Alamos National Laboratory, Los Alamos, NM, 87545 USA e-mail: \{jcastorena, ceodspspectrum, doyen\}@lanl.gov}
}
\maketitle

%%%%%%%%%%%%%%%%%%%%%%%%%%%%%%%%%%%%%%%%%%%%%
%% Abstract
%%%%%%%%%%%%%%%%%%%%%%%%%%%%%%%%%%%%%%%%%%%%%
\begin{abstract}
		{\normalfont
        Binary image based classification and retrieval of documents of an intellectual nature is a very challenging problem. Variations in the binary image generation mechanisms which are subject to the document artisan designer including drawing style, view-point, inclusion of multiple image components are plausible causes for increasing the complexity of the problem. In this work, we propose a method suitable to binary images which bridges some of the successes of deep learning (DL) to alleviate the problems introduced by the aforementioned variations. The method consists on extracting the shape of interest from the binary image and applying a non-Euclidean geometric neural-net architecture to learn the local and global spatial relationships of the shape. Empirical results show that our method is in some sense invariant to the image generation mechanism variations and achieves results outperforming existing methods in a patent image dataset benchmark.
        }
\end{abstract}

%%%%%%%%%%%%%%%%%%%%%%%%%%%%%%%%%%%%%%%%%%%%%
% Sections
%%%%%%%%%%%%%%%%%%%%%%%%%%%%%%%%%%%%%%%%%%%%%

\section{Introduction}
\label{Sec:introduction} 

\begin{figure*}[t] 
	\centering 
	\includegraphics[width=0.8\linewidth]{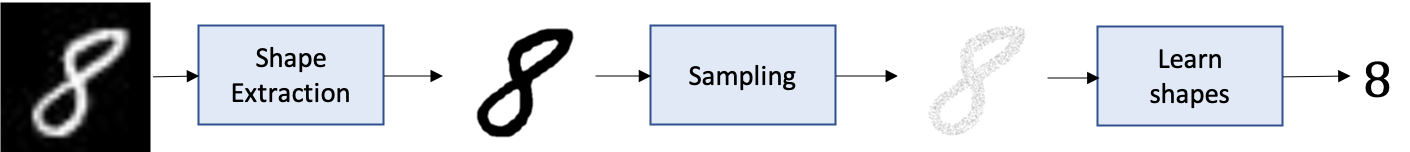}	
	\caption{Image shape analysis system. Shape and background in an image are segmented in the shape extraction block, the shape is then sparsely sampled in the sampling block while the learn shapes block learns features characterizing the spatial relationships between the point samples of the shape.  }
	\label{fig:system} 
\end{figure*}
%

% Application
%Patent based image classification and retrieval is 
Classification and retrieval of documents of intellectual nature is an important and challenging problem. Most widely used methods rely mainly on text and perform the aforementioned tasks by comparing document relevance to text queries. 
%
%Current methods to classify, retrieve and search for documents of intellectual nature rely mostly on text content. Typically, a text query is made and same class or similarity ranked documents are provided for a query request. 
However, intellectual documents typically include and convey valuable information not only through text but also through figures or illustrations. In fact, the USPTO includes in their guidelines for patent application design that "the drawing disclosure is the most important element of the application" \cite{uspto:2020}. Effective methods which exploit such visual information could thus further improve and facilitate the classification, retrieval and search of such documents of intellectual domain. Unfortunately, image based classification/retrieval from intellectual content (e.g., patents, research papers) still remains a very challenging problem. %In this manuscript, attention is focused into applications involving patent documents.
%Image based classification and retrieval of documents of intellectual nature remains a challenging problem. Improving upon image-based methods could facilitate and improve the categorization and search for documents of similar nature. Current, methods for searching exploit  which currently relies for most on text content.  
%The era of information has been shaped in great deal to the powerful text based search engines that classify and retrieve information based on similarity rankings. In the intellectual realm, most search engines produce results based on text queries and have not fully exploited the advantages that could be gained by also analyzing image content. For example, ...talk about patent documents.

%Problems:
Problems affecting effective image based classification and retrieval of documents of intellectual content are  plausibly the image generation mechanisms. These generally include the highly subjective drawing style of the artisan, its preferred view-point to highlight aspects of relevance, the inclusion of multiple objects or components in some sense related and whose spatial distribution may or may not be of importance and the lack of image texture and color. Note here that view-point includes a projection of a 3D rotation and translation into the image reference frame which induces a scaling factor while style includes binary (black and white) images of varying line thickness, dashed/dotted lines and shadings. 

%For example, patent images are generally not rich in texture and are most if not all of the times made up of binary colors. 
%The generation mechanisms 
%%
%\begin{enumerate}
%	\item Binary images are not rich in texture. Should be analyzed in terms of structure or shape rather than on texture.
%	\item View-point variations ( projection of a rigid body transformation (3D rotation and translation) of an object into the 2D image coordinate system) . Here, we assume such projection is random and that we cannot intervene on its generation process. Augmenting is not trivial.
%	\item Drawing style of the patent artist or scientist.
%	\item Presence of multiple components.
%\end{enumerate}
%%

% Potential methods
A number of methods and challenges have been envisioned as attempts to resolve the image based classification/retrieval problem for patent related applications. For example, the work in \cite{Zhiyuan:2007} focuses on retrieval by ranking a Euclidean distance similarity based on the distribution of radial and angular features from image contours. The adaptive hierarchical density histogram (AHDH) method of \cite{Sidiropoulos:2011} consists of an adaptive multi-resolution-multi feature representation with features from black and white pixel counts across the multi-resolution blocks. \cite{Morzinger:2011} fused a set of local handcrafted features including local binary patterns and edge histograms to train a linear SVM. The last relevant work achieving the highest accuracy is the work of \cite{Csurka:2011}. This method, constructed Fisher vectors (FV) through the parametric estimation of a Gaussian mixture model (GMM) representation of the distribution of local SIFT features computed across the entire image. Distinct classifiers were trained with SVM significantly outperforming the others.
%
%A number of methods and challenges have been envisioned as attempts to resolve the image based classification/retrieval problem from images of intellectual content. For example, the work in \cite{Morzinger:2011} fused a set of local handcrafted features including local binary patterns and edge histograms to train a linear SVM achieving an accuracy of 66\%. \cite{Csurka:2011} constructed Fisher vectors (FV) through the parametric estimation of a Gaussian mixture model (GMM) representation of the distribution of local SIFT features computed across the entire image. Distinct classifiers were trained with SVM significantly outperforming the others. The method of \cite{Zhiyuan:2007} focuses on retrieval by ranking a Euclidean distance similarity based on the distribution of radial and angular features from image contours. The last relevant work we include here is the adaptive hierarchical density histogram (AHDH) method of \cite{Sidiropoulos:2011} that consists of an adaptive multi-resolution-multi feature representation where features are based on black and white pixel counts across the multi-resolution blocks. 
%
Unfortunately, most of these methods suffer in performance achieving classification accuracies around 65\% with the exception of \cite{Csurka:2011} with  90\%.  % for since they have been drawn from the generic and vast image analysis literature without considering the image generation mechanisms causing variations.

% A potential solution CNN's
Convolutional neural networks (CNN's) \cite{Lecun:98} within deep learning (DL) have revolutionized the way by which we do image analysis and have outperformed traditional methods in many application problems \cite{LeCun:2015}. Connecting such research advances to solve the problems being faced for the classification/retrieval of images of intellectual property would be highly benefitial. However, application of off-the-shelve CNN architectures operating in the Euclidean image domain although straightforward are not the most efficient and do not exploit the full potential of CNN's for this specific problem mainly because binary images are texture-less. %But secondly, because there is a functional dependence $p(y|x)$ on the data distribution $p(x)$ which makes it invariant to the patent image generation mechanisms. Augmenting datasets with sufficient examples characterizing the generation variations could alleviate the problem
But secondly, because of their inability to cope with the generation mechanisms causing variations in the absence of sufficient data characterizing all potential combinations of these \cite{Parascandolo:2018}. This later, i.e., controlling and intervening the highly subjective nature of the generation process of intellectual images as well as its access to generate a sufficient dataset is extremely challenging. %{\color{red} include here a reference }

%imagepushed forward previously achieved performances in many applications involving images, audio, and generic signals laying on a d-dimensional Euclidean domain. The driving engine of such advances are convolutional neural networks (CNN's) \cite{Lecun:98} that exploit the richness of texture information on images to learn shift-invariant features characterizing the image content. Some of the main properties of such architectures are the ability to learn the fundamental features across different scales, its computational efficiency due to the Fourier convolutional theorem, the reduced number of parameters required by CNN's in comparison to fully-connected layers and finally the flexibility to parallelize algorithms. 

% Problems in using CNN's

% The approach we seek  (Problem formalization or formulation)
Here, we propose a learning based method that aims both at making the application of neural-nets on binary images more efficient and at disentangling the learned functionals from some of the variations in the image generation mechanisms.  %making it thus more robust to such transformations. 
Efficient application of the neural net is made by operating directly on the object or shape of interest contained and extracted from the patent image rather than on the entire image. This of-course implies usage of a neural-net operating on the non-Euclidean domain learning local and global features characterizing the spatial relationships of the shape. Operating on the shape allows one to disentangle from the generation mechanisms of scaling, translation by centering and normalizing while plausibly introducing some invariance to rotations and style from the learned local and global features characterizing spatial relationships. %Finally, we 
%
%
% Summary of our contribution
%Our contribution in this research is a method combining the extraction of sparse point clouds from image shapes and learning local and global spatial relationships between the points that characterize shapes. We contend that learning graph node and edge relationships can aid in such characterization and propose the use of the dynamic graph CNN (DGCNN) of \cite{Wang:2019}. Such an architecture dynamically applies convolution like operators to the non-Euclidean domain and has capabilities to learn both local and global features diffusing non-locally across layers for improved inference. %This later advantage is derived from the property that DGCNN can dynamically generate different graphs at each layer thus adding a capability for the non-local diffusion of features. 
Given this, our paper presents the proposed approach in Section~\ref{Sec:approach}. Section~\ref{Sec:experiments} presents experimental results comparing classification/retrieval performance across different datasets with scaling, rotation and translation transformations, and finally Section~\ref{Sec:conclusion} concludes our findings.

						% Introduction
\section{Approach}
\label{Sec:approach}

% General overview of the method
The basis of the method we propose consists on the extraction of shapes from images, point sampling from the shape and finally learning node and edge interconnection features that characterize shape point samples. A summary of the system is shown in Figure \ref{fig:system}. %Below we detail the mechanisms by which these three steps are accomplished. 

\begin{figure*} 
	\centering 
	\includegraphics[width=1.0\linewidth]{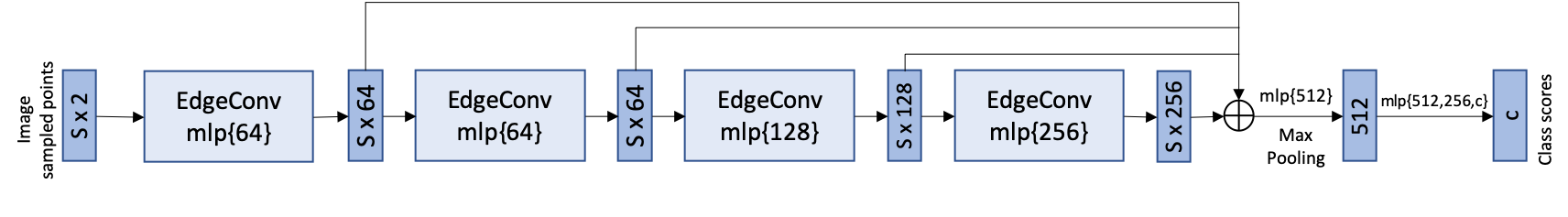}	
	\caption{Dynamic Graph CNN Architecture.  }
	\label{fig:architecture} 
\end{figure*}
\begin{figure} 
	\centering 
	\includegraphics[width=0.5\linewidth]{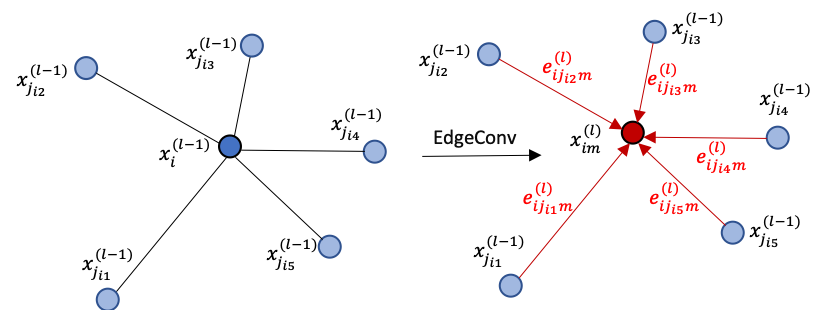}	
	\caption{EdgeConv at a node.  }
	\label{fig:edgeConv} 
\end{figure}
%%Below we detail the mechanisms by which these three steps are accomplished. 

\subsection{ Point clouds from image shapes }
\label{Ssec:pc}
Shapes are first extracted from images by segmenting shape from background through the well known adaptive thresholding method of Otsu \cite{Otsu:1979}. Note that the use of a shape segmentation method is a necessity only in cases when images are non-binary. Such an extraction and analysis on the non-Euclidean spatial composition of a shape instead of the Euclidean analysis on image pixel values will enable the learning engine to be invariant against image texture or color; a property of crucial importance to many applications.
%Representing image content through shapes (i.e., spatial and geometrical relationships) introduces invariance to image texture; a property of crucial importance to the aforementioned applications of relevance.

Following, a representative sparse point-cloud of the shape is extracted by clustering the pixel locations conforming the segmented shape.  The algorithm we use here for such clustering is the mini-batch $k$-means \cite{Sculley:2010} which is orders of magnitude more computationally efficient than standard $k$-means with only small penalties in performance. The efficiency of such method is obtained by processing data iteratively in batches of randomly picked points with centroids computed through stochastic gradient descent (SGD). Each of the estimated $S$ centroids denoted here by $\x^{(0)}_i \in \Rspace^2$ for $i \in \{1,..,S\}$ contains the corresponding $(x,y)$ location of the estimates. Comforming all of the $k$-means centroids results in a point cloud $\X^{(0)} = \{\x^{(0)}_1,..., \x^{(0)}_S \}$ of sparsely sampled representative points of the shape. Note that, throughout this paper the superscript $^{(0)}$ denotes an input. 

Shape sampling is solely meant here to reduce the computational complexity of the learning algorithm described in the next section, specially in cases when the shape is not complex. However, dense point clouds including all the $(x,y)$ pixel locations conforming a shape could also be used instead. This of course at the cost of higher computational complexity imposed on the learning algorithm.

\subsection{ Learning shapes through graphs }
\label{Ssec:learning_graphs}
% Driving idea of dynamic graph CNNs: 
The sampled points $\X^{(0)}$ of the shape are used as input to a deep neural net that learns features characterizing the meaningful spatial relationships between the points. Here, the sampled points and their spatial relationships are represented as nodes and edges of a graph, respectively. The learning engine of such graphs is a dynamic graph CNN (DGCNN) \cite{Wang:2019} architecture dynamically operating and learning graphs across its layers. Here, each layer $l$ produces graphs $\G^{(l)}_m = (\V^{(l)}, \E^{(l)}_m)$ indexed by $m \in \{1,..,M^{(l)}\}$ by dynamically associating $k$ nearest learned feature neighbors (f-$k$NN's) according to node distances in the feature space. %Such property allows the network to diffuse features non-locally
% Description of components of the network: EdgeConv network performs 
An instance of the DGCNN architecture we use is shown in Figure \ref{fig:architecture}. Such an architecture consists of a sequence of $M^{(l)}$-dim edgeConv layers consisting each of node and edge connection features $\x^{(l)}_i \in \Rspace^{M^{(l)}}$ for $i \in \V^{(l)} $ and $\e^{(l)}_{ij} \in \Rspace^{M^{(l)}}$ for $(i,j) \in \E^{(l)}_m$, respectively.  Node features $x^{(l)}_{im}$ are computed element-wise as 
\begin{equation} \label{featureEq}
x^{(l)}_{im} = \max\limits_{j:(i,j) \in \E_m^{(l)}} e^{(l)}_{ijm}
\end{equation}
where the maximum is computed over the f-$k$NN's at the node $i$ and edge features computed element-wise as  
\begin{equation} \label{edgeEq}
e^{(l)}_{ijm}= \text{ReLU} ( \th^{(l)}_m \cdot ( \x^{(l-1)}_j - \x^{(l-1)}_i) + \ph^{(l)}_m \cdot \x^{(l-1)}_i)
\end{equation}
with $\th^{(l)}_m, \ph^{(l)}_m \in \Rspace^{M^{(l)} }$ being parameters of the learned convolution filters. Note that the first inner product in Eq \eqref{edgeEq} captures local information whereas the second inner product captures global information. Fig. \ref{fig:edgeConv} illustrates the mechanism by which the edgeConv layer generates features in a node. First it computes the f-$k$NNs $j_{i_1},..., j_{i_5} $ centered at node $i$ using Euclidean distances between features. Then, computes functions $e'_{ijm}$ in Eq. \eqref{edgeEq} through the learned filters only for the f-$k$NNs and then applies \eqref{featureEq}. Implementations of these operators can be framed as a shared multi-layer perceptron (mlp).						% Approach 				
\section{Experiments} 
\label{Sec:experiments}
%

% MNIST dataset and classification task
Evaluation of the proposed approach is tested against the task of classification of the MNIST benchmark \cite{Lecun:98} and the CLEF-IP dataset of patent images in \cite{Piroi:2011}. %The datasets are partitioned into balanced-class training and testing subsets. 
The intent to include experiments on the MNIST dataset is to demonstrate the inability of standard CNN's operating in the Euclidean domain to cope with data transformations in the abscence of sufficient data characterizing these. A matter of crucial importance to build effective classification/retrieval solutions in the patent application realm.

% Point cloud representation: Include a few illustrations of the point cloud representation of MNIST digits
\subsection{ Point cloud extraction } \label{SSec: SP}
\begin{figure} [htb]
	\centering 
	
	\subfloat{\includegraphics[width=0.09\linewidth]{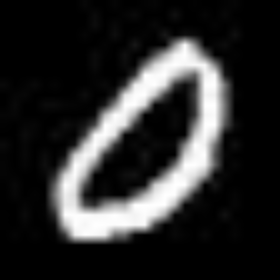}}
	\subfloat{\includegraphics[width=0.09\linewidth]{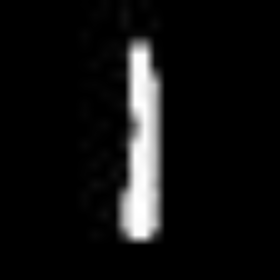}}
	\subfloat{\includegraphics[width=0.09\linewidth]{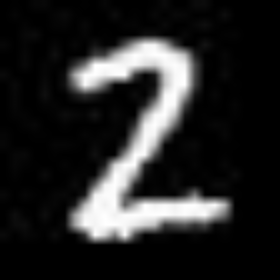}}
	\subfloat{\includegraphics[width=0.09\linewidth]{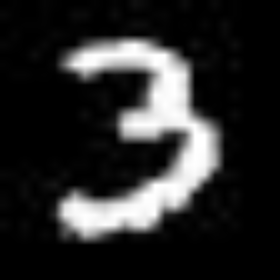}}
	\subfloat{\includegraphics[width=0.09\linewidth]{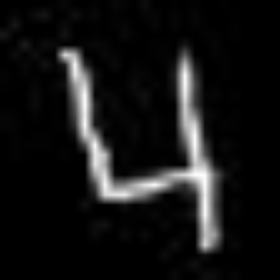}}
	\subfloat{\includegraphics[width=0.09\linewidth]{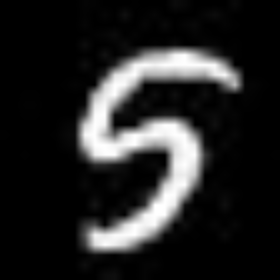}}
	\subfloat{\includegraphics[width=0.09\linewidth]{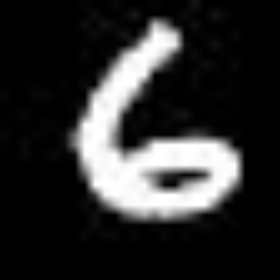}}
	\subfloat{\includegraphics[width=0.09\linewidth]{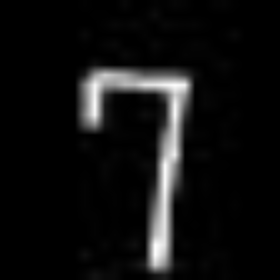}}
	\subfloat{\includegraphics[width=0.09\linewidth]{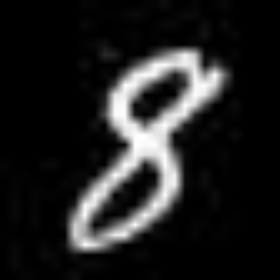}}
	\subfloat{\includegraphics[width=0.09\linewidth]{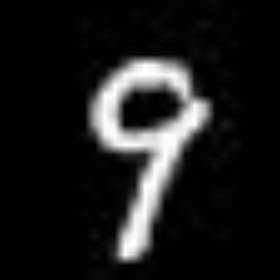}}	
	
	\vspace{-0.9em}
	
	\subfloat{\includegraphics[width=0.09\linewidth]{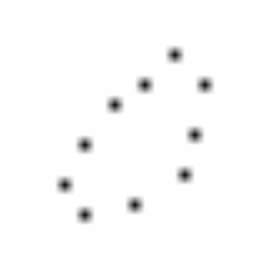}}
	\subfloat{\includegraphics[width=0.09\linewidth]{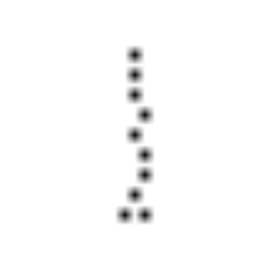}}
	\subfloat{\includegraphics[width=0.09\linewidth]{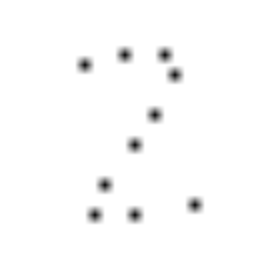}}
	\subfloat{\includegraphics[width=0.09\linewidth]{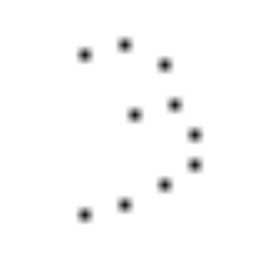}}
	\subfloat{\includegraphics[width=0.09\linewidth]{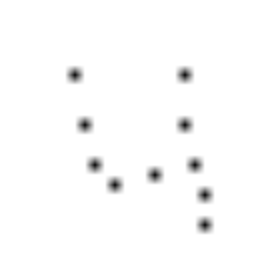}}
	\subfloat{\includegraphics[width=0.09\linewidth]{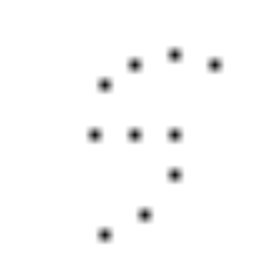}}
	\subfloat{\includegraphics[width=0.09\linewidth]{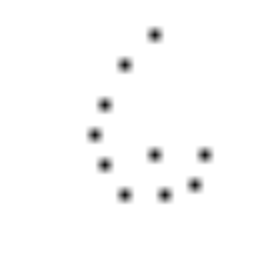}}
	\subfloat{\includegraphics[width=0.09\linewidth]{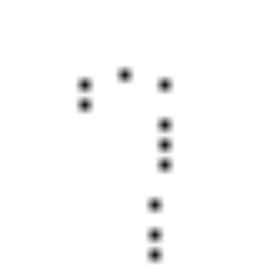}}
	\subfloat{\includegraphics[width=0.09\linewidth]{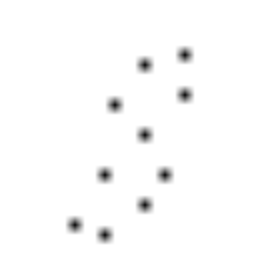}}
	\subfloat{\includegraphics[width=0.09\linewidth]{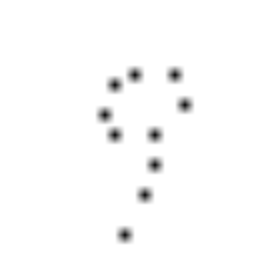}}	

	\caption{Sampling points from MNIST digit images. Row~1: MNIST images, Row~2: Sampled points.}
	
	\label{fig:superpixels} 
\end{figure}

\begin{figure}[htb]
	\centering 
	\subfloat[{\tiny MNIST}]{\includegraphics[width=0.11\linewidth]{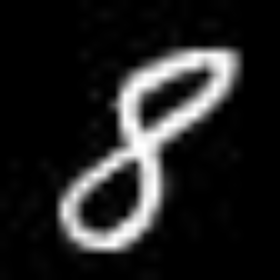}}
	\subfloat[{\tiny S=35}]{\includegraphics[width=0.11\linewidth]{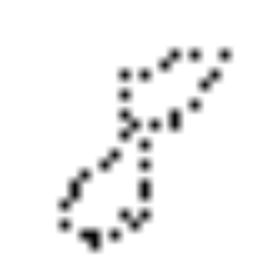}}
	\subfloat[{\tiny S=30}]{\includegraphics[width=0.11\linewidth]{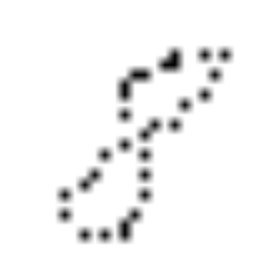}}
	\subfloat[{\tiny S=25}]{\includegraphics[width=0.11\linewidth]{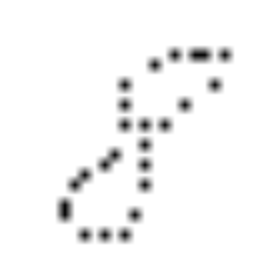}}
	\subfloat[{\tiny S=20}]{\includegraphics[width=0.11\linewidth]{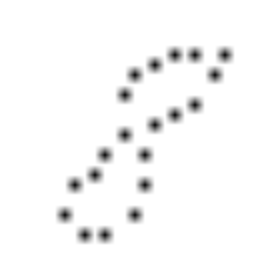}}
	\subfloat[{\tiny S=15}]{\includegraphics[width=0.11\linewidth]{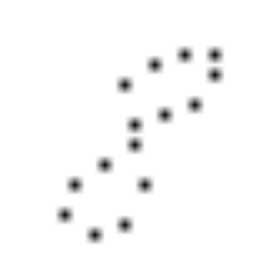}}
	\subfloat[{\tiny S=10}]{\includegraphics[width=0.11\linewidth]{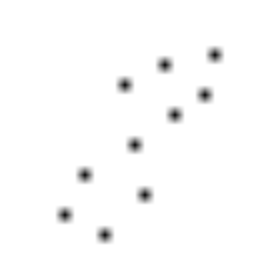}}
	\subfloat[{\tiny S=5}]{\includegraphics[width=0.11\linewidth]{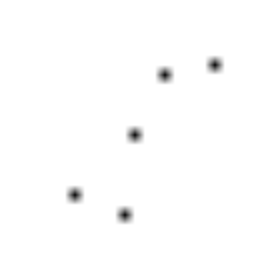}}
	\caption{MNIST digit point cloud as a function of sparsity}
	
	\label{fig:sp_reduction} 
\end{figure}

\begin{figure}[h] 
	\centering 
	\subfloat[Bike seat image $2254 \times 1854$ ]{\includegraphics[width=0.25\linewidth]{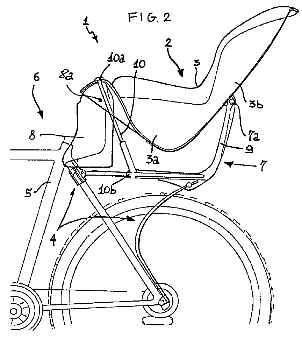}}
	\subfloat[Circuit image $920 \times 1248$]{\includegraphics[width=0.25\linewidth]{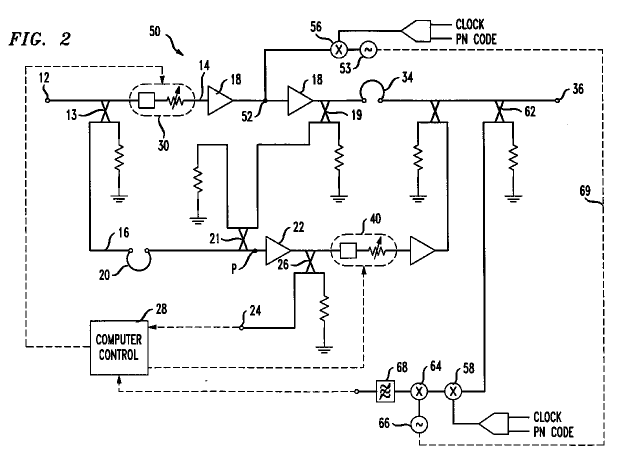}}	
	\subfloat[Helicopter image $964 \times 1440$ ]{\includegraphics[width=0.25\linewidth]{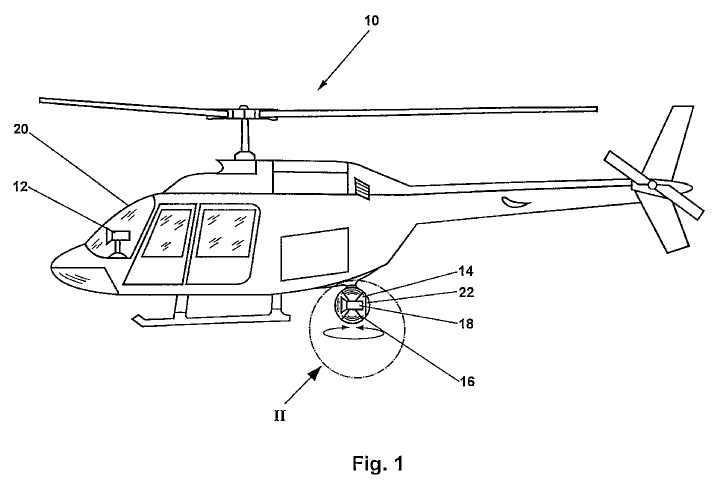}}

	\subfloat[Sampled image shape]{\includegraphics[width=0.25\linewidth]{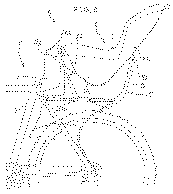}}
	\subfloat[Sampled image shape]{\includegraphics[width=0.25\linewidth]{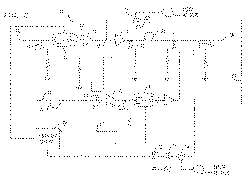}}
	\subfloat[Sampled image shape]{\includegraphics[width=0.25\linewidth]{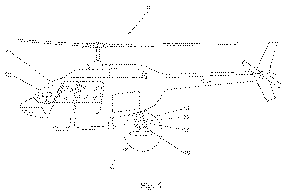}}
	
	\caption{Sparse sampling of CLEF-IP image shapes.  }
	\label{fig:patents} 
\end{figure}

We generate a point-cloud for each image shape in the aforementioned datasets following Section \ref{Ssec:pc}. 
% MNIST
Figure \ref{fig:superpixels} illustrates a few representative point cloud examples where the first row shows the MNIST digit images and the second row represents the corresponding extracted point clouds. The point-clouds are each of size $10 \times 2$ (i.e., $S=10$ ) for which case the mini-batched $k$-means algorithm runs near real-time and thus does not represent a significant computational bottleneck.
Here, we also observe that the proposed sampling mechanism captures the underlying structure of the MNIST shapes even when $S=10$. For illustration purposes we also include in Figure \ref{fig:sp_reduction} an example that shows the effect of increasing sparsity. Note that at the sparsest point cloud it is visually difficult to identify the digit and some ambiguity exists as to whether it is a 2, 3 or an 8. Fortunately, with sparsity of at least $S=10$ the digit can be identified. Such effect is consistent with at least a vast number of examples manually inspected from the MNIST. %justifies the usage of an algorithm like the one we use in place of something simpler like random sampling which may not capture or preserve the shape structure.
In addition, we also include representative examples of the sampled shapes extracted from CLEF-IP's \cite{Piroi:2011} images extracted from patent documents. Figs.~\ref{fig:patents}(a-c) are the raw images while Figs.~\ref{fig:patents}(d-f) are the results of sparsely sampling the corresponding image shapes in (a-c) with $S=1000$. Note that the sampling mechanism we employ visually preserves the shapes in all three cases as long as the point-cloud is relatively dense.

\subsection{Classification} \label{SSec: DGCNN}

% Describe the network architecture used
Following the architecture model of \cite{Wang:2019} we use four EdgeConv layers as shown in Figure \ref{fig:architecture} of which the first three are fully-connected producing, respectively, features of size (64, 64, 128, 256). Max/sum pooling is used, unless otherwise specified the number of f-$k$NN is set to 5, dropout with keep probability of 0.5 and all layers include LeakyReLU and batch normalization. 

% Describe the parameters used for training
For training, we use stochastic gradient descent (SGD) with self-adaptive learning rate. The rule to self-adapt is based on the warm restart method of \cite{Loshchilov:2016} with an initial value of 0.1. Batch size is set to 32, momentum for batch normalization is set to 0.9 and \#epochs = 100. We tested performance with a variety of number of parameters and the ones selected are the ones that yielded the best results.

% Describe the data set used for training
Shapes are extracted from the image datasets by first transforming them into binary images through segmentation with Otsu's adaptive thresholding method \cite{Otsu:1979}. After the shape-background segmentation, each shape is sampled following the method description in Section \ref{Ssec:pc}. A point cloud of size $S \times 2$ is obtained for each $ 28 \times 28$ image that is subsequently centered and normalized to be within the unit $\ell_2$-norm ball.

\begin{figure}[h]
	\centering 
	\includegraphics[width=0.5\linewidth]{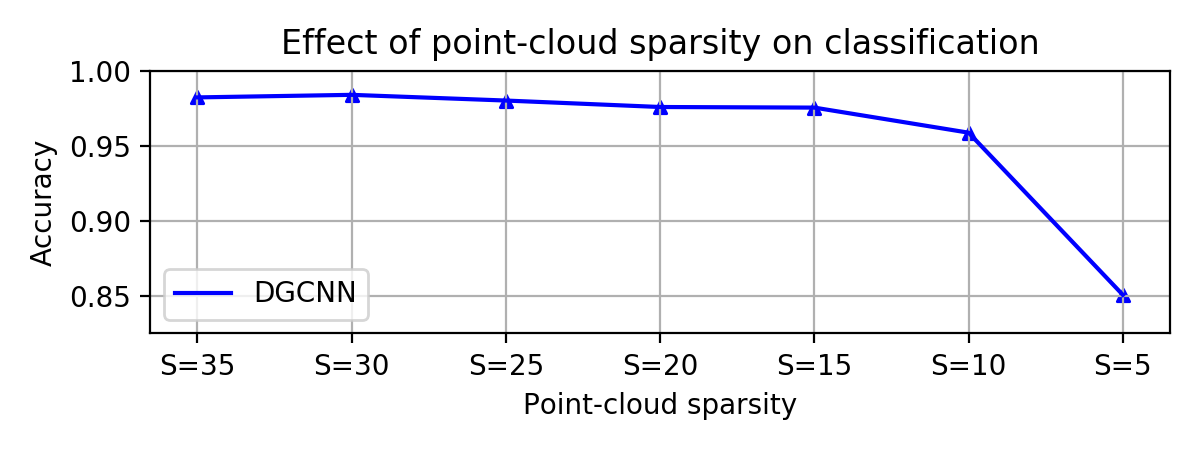}
	\caption{Classification versus point cloud sparsity.}
	\label{fig:sparsity_performance} 
\end{figure}

%%% Results
%% MNIST
% Test against point cloud sparsity
The first experiment is intended to obtain MNIST classification performance against the sparsity of the point cloud extracted from the image shapes. This, to justify sampling a shape sparsely with higher computational complexity benefits over denser points clouds. For this experiment, partitions of 42K and 28K images were utilized to train and test, correspondingly. Figure \ref{fig:sparsity_performance} illustrates classification performance as a function of point cloud sparsity. Our results validate the intuition that classification performance does not suffer with sparser point-clouds above a certain level; in this case $S>15$. At sparsity $S=5$ the shape is no longer visually identifiable from other sampled shapes as observed in Figure \ref{fig:sp_reduction}.h. %Note here that there is no gain in classification performance while higher computational complexities in learning are imposed with denser point clouds $S>20$ for the MNIST shape complexity.
%
%We also note that the DGCNN architecture achieves good classification accuracy by using only point clouds of size $10 \times 2$ from the image which in itself sounds very promising for learning more complex shapes such as those present in documents of intellectual property (e.g., patent drawings, diagrams) or scientific documents (e.g., plots, equations).

\begin{figure}[h]
	\centering 
	\includegraphics[width=0.5\linewidth]{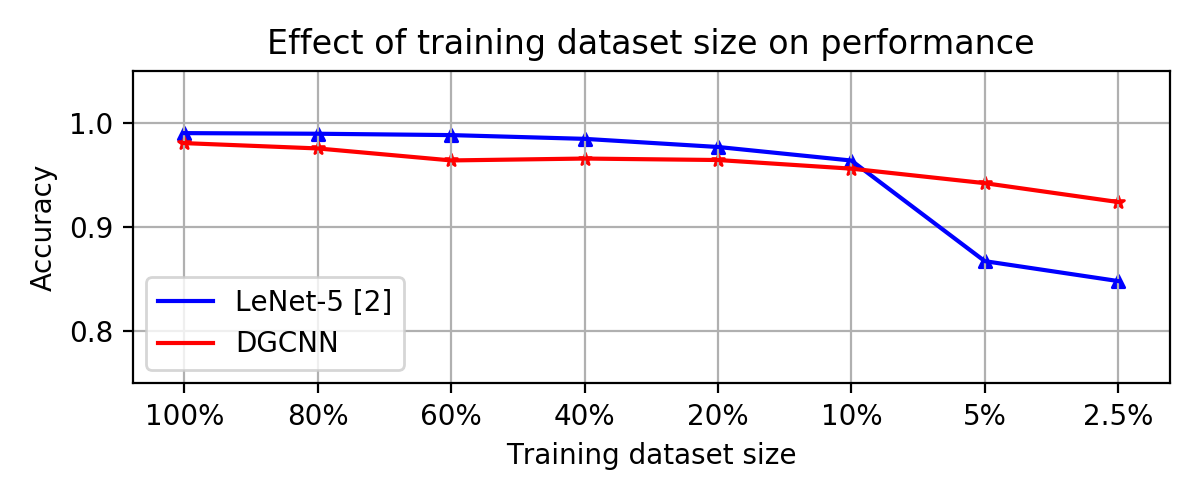}
	\caption{Classification versus training dataset size.}
	\label{fig:generalization} 
\end{figure}

% Tests to compare againsts other methods
A comparison of MNIST classification performance against a CNN method as training dataset size is varied was also conducted with results included in Fig. \ref{fig:generalization}. This with the intent to observe the capabilities of the neural-net to learn features effectively when dataset size is reduced significantly. The methods we compared against are the LeNet-5 \cite{Lecun:98} with full $28 \times 28$ image inputs and the DGCNN architecture. Note that the LeNet-5 network can be considered as one of the smallest sized neural-nets out there thus being one of the least data hungry architectures which is of high appeal for our application. %We observe that the DGCNN model consistently outperforms PointNet at all training dataset sizes. This indicates that the addition of modules to learn global features in the DGCNN improves shape characterizations over PointNet learning only local shape features. 
When comparing against LeNet-5, our findings show that the DGCNN performs consistently with variations in dataset size even better than the LeNet-5 in the lower size extreme case. 

Evaluations on classification were performed now against the effect of data transformations that can plausibly influence the patent image generation mechanisms.  Two experimental cases are included with the transformed test data: (1) when training examples include images subjected to the corresponding transformations and (2) when training does not include examples of the transformed data. Since the MNIST dataset is the simplest to manipulate for transformations we use it as our comparison benchmark. The transformations for which we test here are random uniformly sampled from the limit $\pm 90^{\circ}$ for rotations, $\pm$ 9 pixels for translations in both horizontal and vertical directions and scaling in the (0.2,1) level. Table \ref{tab:table2} summarizes the comparison results of performance against such transformations. Here, the results at the same row as the transformation label (e.g., "scale") show performance in the case of (2) i.e., when training data does not include transformation examples.  The row where the "w re-training" label is (right under the transformation label) corresponds to performance of (1) when transformation examples of the expected transformation levels are included in the training data. 
%
%Each transformation applied independently. Second, we train and test performance against the three transformations applied altogether to the MNIST dataset. In such case, rotations, translations and scaling are randomly sampled from a uniform distribution with limits $\pm 90^{\circ}$, $\pm$ 9 pixels and 0.3 for rotation, translation and scaling, respectively.
%Table \ref{tab:table2} summarizes the comparison results of performance against transformations of the MNIST dataset.
%% 
%\begin{table}[htb]
%	\caption{\label{tab:table2} Classification against MNIST transformations.}
%	\centering
%	\begin{tabular}{lcccc}
%		\hline
%		%\\
%		\multirow{3}{*}{Method} & \multirow{2}{*}{scaled} & \multirow{2}{*}{rot} & \multirow{2}{*}{trans} & scaled \\
%		&  &  & & rot, trans\\
%		& MNIST & MNIST & MNIST & MNIST\\
%		\hline
%		\\
%		LeNet5\cite{Lecun:98} & 0.961 & \pmb{0.981} & 0.976 & 0.874\\
%		Proposed & \pmb{0.985} & 0.933 & \pmb{0.995} & \pmb{0.935}\\
%		\hline
%	\end{tabular}
%\end{table}
%%
% 
\begin{table}[htb]
	\caption{\label{tab:table2} Classification against transformations on the MNIST dataset.}
	\centering
	%\resizebox{1.0\columnwidth}{!}{
	\begin{tabular}{lcc}
		\hline
		\\
		Transformation & LeNet-5 \cite{Lecun:98}  & Proposed \\
		\\
		\hline
		\\
		scale  & 0.630 & \pmb{0.985} \\
		\quad {\small w re-training}  & 0.961 & \\
		\hline
		rotation & 0.562 & 0.933 \\
		\quad {\small w re-training}  & \pmb{0.981} &  \\
		\hline 
		translation &  0.286 & \pmb{0.995}\\
		\quad {\small w re-training}  & 0.976 &  \\
		\hline
		scale, rot & \multirow{2}{*}{0.134} & \multirow{2}{*}{\pmb{0.935}} \\
		translate & &  \\
		\quad {\small w re-training}  & 0.874 &  \\
		\hline 
		binary inversion & 0.257 & \pmb{0.992} \\
		\quad {\small w re-training}  & 0.991 &  \\
		\hline
	\end{tabular}
	%	}
\end{table}
These results show that our proposed framework remains more or less invariant to the aforementioned transformations in comparison to LeNet-5 and that it outperforms it in most cases in its own intended benchmark dataset. The centering and $\ell_2$ normalization of point clouds from extracted shapes makes our method invariant to scaling and translation transformations while some invariance to rotations is learned by the neural net exploiting spatial relationships between sampled points. In general, we can say that analysis of shapes extracted from images with the proposed method better copes to rigid body transformations in the 2D plane and across scales. However, this is not necessarily true for general view-point 3D-2D projections as these may include overlapping but potentially different information depending on the view-point drawing mechanism. The result in the binary inversion case was interesting, it showed that LeNet-5 suffered when the background was negatively correlated with the expected shape pixel level, a performance degradation issue mentioned in \cite{Schlkopf:2019}.

\begin{figure} [htb]
	\centering 
	
	\subfloat[Drawing]{\includegraphics[width=0.2\linewidth]{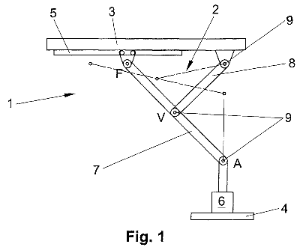}}	
	\subfloat[flowchart]{\includegraphics[width=0.2\linewidth]{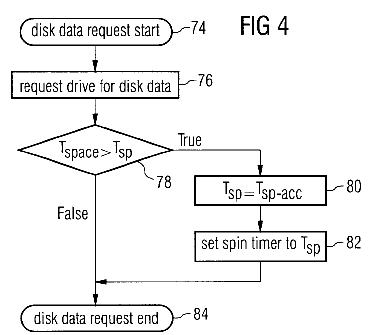}}
	\subfloat[Graph]{\includegraphics[width=0.2\linewidth]{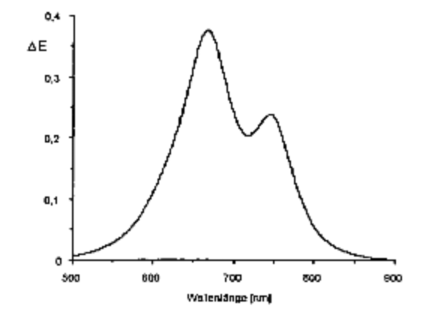}}
	\subfloat[Symbol]{\includegraphics[width=0.2\linewidth]{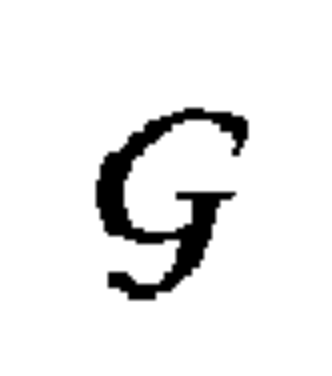}}	
	\subfloat[Program]{\includegraphics[width=0.2\linewidth]{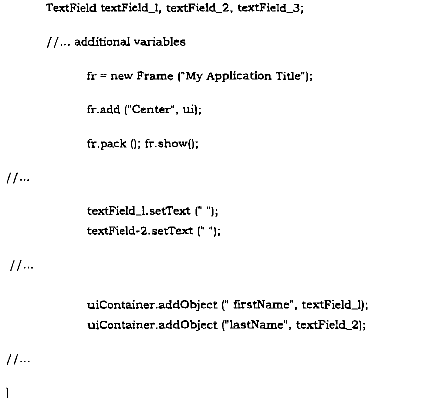}}	
	
	\subfloat[Table]{\includegraphics[width=0.23\linewidth]{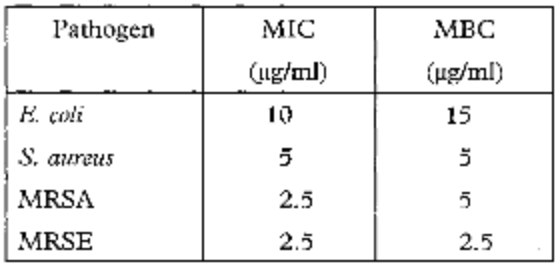}}		
	\subfloat[Math]{\includegraphics[width=0.23\linewidth]{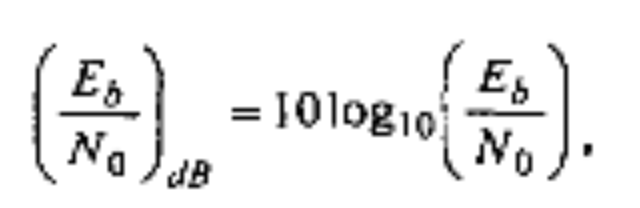}}	
	\subfloat[Chem]{\includegraphics[width=0.23\linewidth]{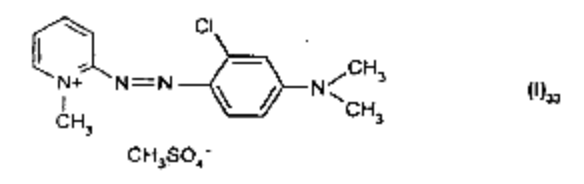}}
	\subfloat[Geneseq]{\includegraphics[width=0.23\linewidth]{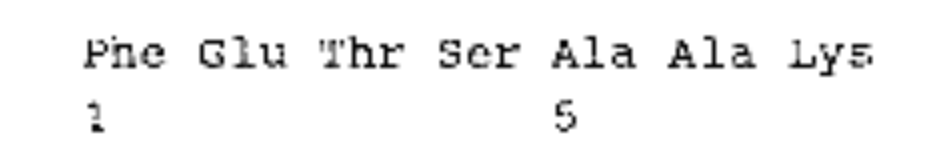}}
	\caption{Examples of the CLEF-IP 9-class images.}
	
	\label{fig:clefip_images} 
\end{figure}

\begin{figure} [htb]
	\centering 
	
	\subfloat[Top-side-view]{\includegraphics[width=0.33\linewidth]{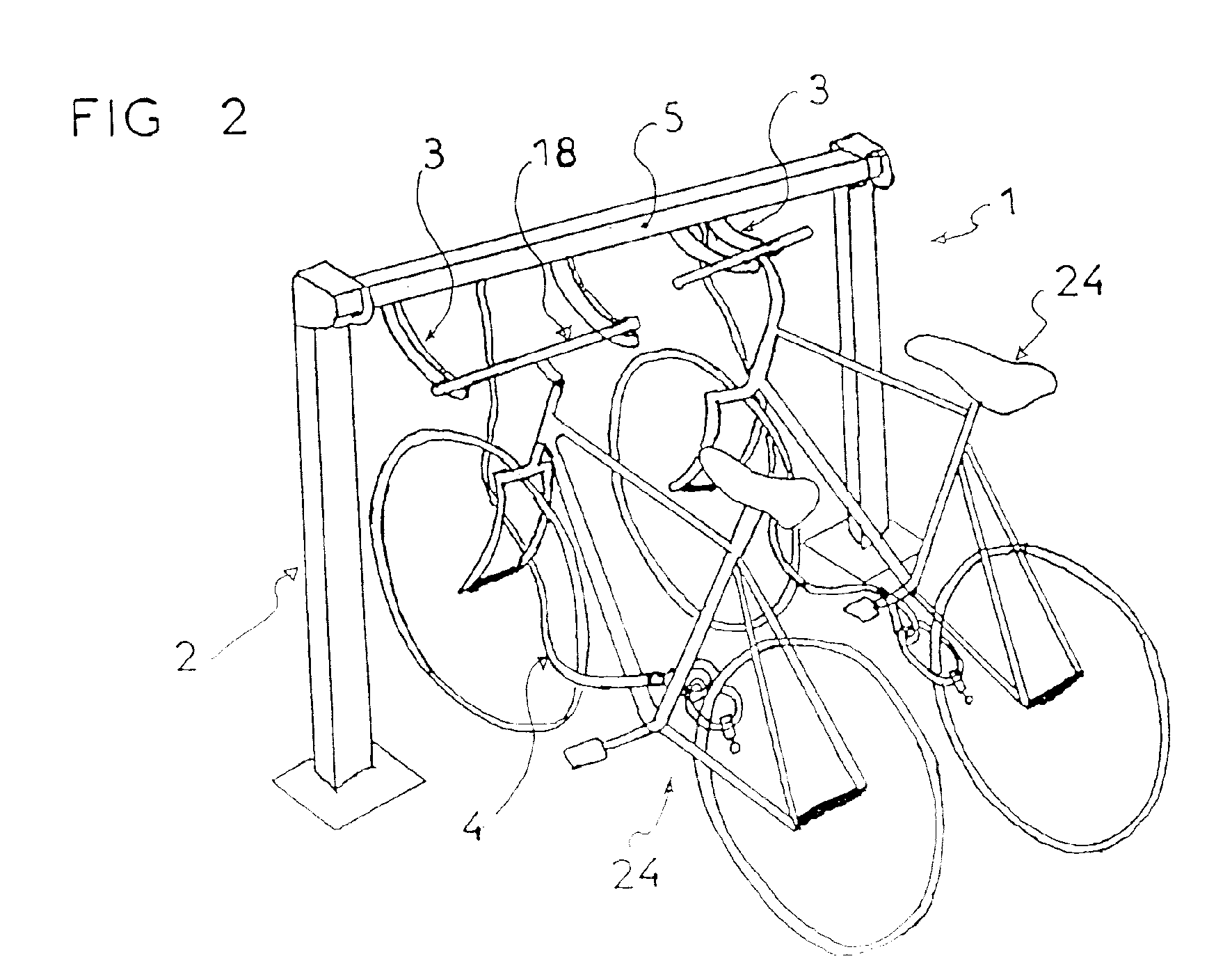}}
	\subfloat[Side-view]{\includegraphics[width=0.33\linewidth]{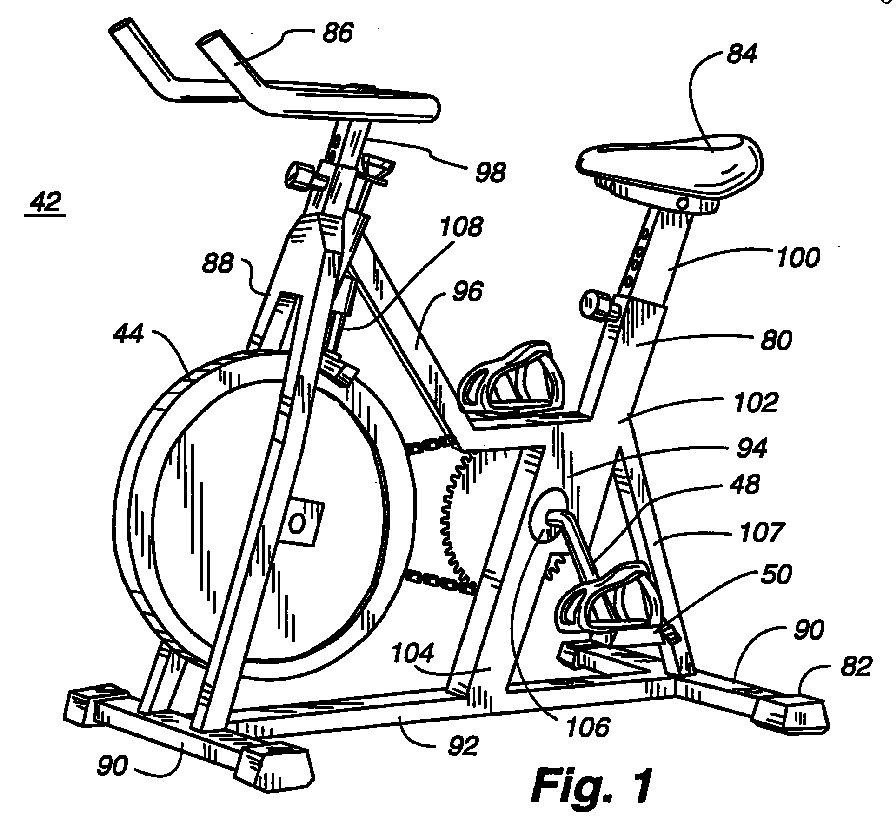}}
	\subfloat[Cross-class components]{\includegraphics[width=0.23\linewidth]{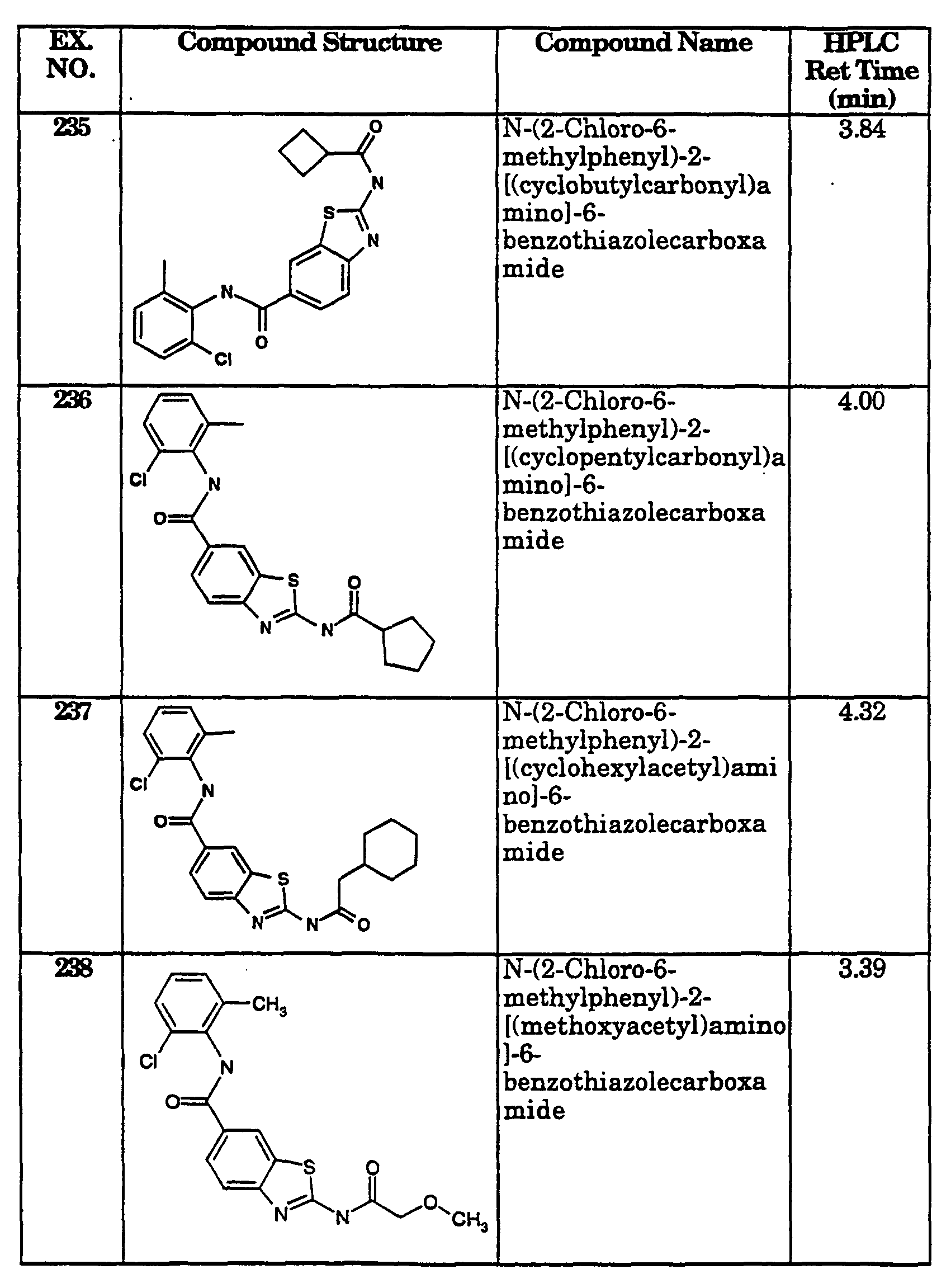}}
	\caption{Variations of CLEF-IP patent figures.}
	
	\label{fig:clefip_var} 
\end{figure}

The final experiment on classification evaluates the proposed method against the CLEF-IP \cite{Piroi:2011} patent image dataset. This dataset was built as a classification challenge including image categorization into one of the 9 classes: drawing, flowchart, graph, symbol, math, table, program, chem, geneseq. A representative example of the images for each of these classes is shown in Figure \ref{fig:clefip_images}. 
Note that CLEF-IP dataset contains image shapes from an unknown and varying viewpoint. Such viewpoints are assumed here to be a projection of a rigid body transformation (3D rotation and translation) into the 2D image space. Although such viewpoints and generation process of patent figures are decided at the best discretion of the patent artist to highlight certain aspects of the inventions, we assume here that this mechanism is random and that we have no intervention power on its generation. In addition, some figures contain multiple cross-class components which represents more challenges for learning class characteristics. Figure \ref{fig:clefip_var} illustrates examples of some of these variations together with an example presenting multiple components in a patent image.

The CLEF-IP dataset contains approximately 38K unbalanced-class images partitioned into the training and testing subsets. Each image was sampled and used as input to a DGCNN to learn the spatial relationships characterizing the shape classes. The sparsity of sampling was set in this case to $S=1000$ and the f-$k$NN was set to 25. Intuition behind the usage of such sparsity is that in general the CLEF-IP contains images of significantly higher complexity compared to the simpler MNIST shapes. To learn the spatial relationships between the shape samples, the hyperparameters of the DGCNN were set to be the same as those described at the beginning of this section. For comparison, we also include classification performance against two standard CNN methods: the basic LeNet-5 and the more sophisticated residual image learning ResNet-50 architecture \cite{He:2016}. In addition, we include results from \cite{Csurka:2011} whose method was specifically designed for patent image based classification and whose performance is the best for this task, to the best of our knowledge. The results we obtained were summarized in table \ref{tab:table3}. 

\begin{table} [htb]
	\caption{\label{tab:table3} Classification performance on CLEF-IP \cite{Piroi:2011}}.
	\centering
	\resizebox{0.65\columnwidth}{!}{
		\begin{tabular}{lcccc}
			\hline
			\multirow{2}{*}{} & \multirow{2}{*}{LeNet-5\cite{Lecun:98}} &  \multirow{2}{*}{Resnet-50 \cite{He:2016}} & FV & \multirow{2}{*}{Proposed}  \\
			& &  & + SVM \cite{Csurka:2011} & \\
			\hline
			\\
			& 0.555 & 0.244 & 0.907 & \pmb{0.942}\\
			\hline
		\end{tabular}
	}
\end{table}
Empirical results reflect that CNN's operating in the Euclidean domain are not powerful enough to disentangle performance from view-point, drawing style and the presence of multiple components in binary patent imagery. The method of \cite{Csurka:2011} based on the Fisher vector representation performs nicely and better than the standard CNN methods for this task. However, our proposed method seems to better cope with the plausible variations in the image generation mechanism and outperforms all compared methods which includes to the best of our knowledge the best so far for the specific task.

\subsection{ Retrieval} \label{SSec: DGCNN-retr}
In addition to the classification task we also compare the performance of the proposed approach against the standard structural similarity Index (SSIM) method \cite{hore2010image} in a retrieval task. We tested against the MNIST, fashion-MNIST \cite{Xiao:2017} and CLEF-IP datasets described in the previous subsection and  show that the proposed framework performs significantly better than the SSIM approach. To measure performance, we use the mean average precision (MAP) overall retrievals computed as:
\begin{equation}
\text{MAP} = \frac{\sum_{q=1}^{Q} \text{Ave} (P(q)) } {Q}
\end{equation}
where $Q$ is the number of queries and $\text{Ave} (P(q))$ is the precision average of score for each query $q$. Here, $\text{Ave} (P(q))$ is obtained by computing the average number of items correctly retrieved in a $k$ nearest neighborhood.

\begin{figure}[htb]
	\centering 
	\includegraphics[width=0.65\linewidth]{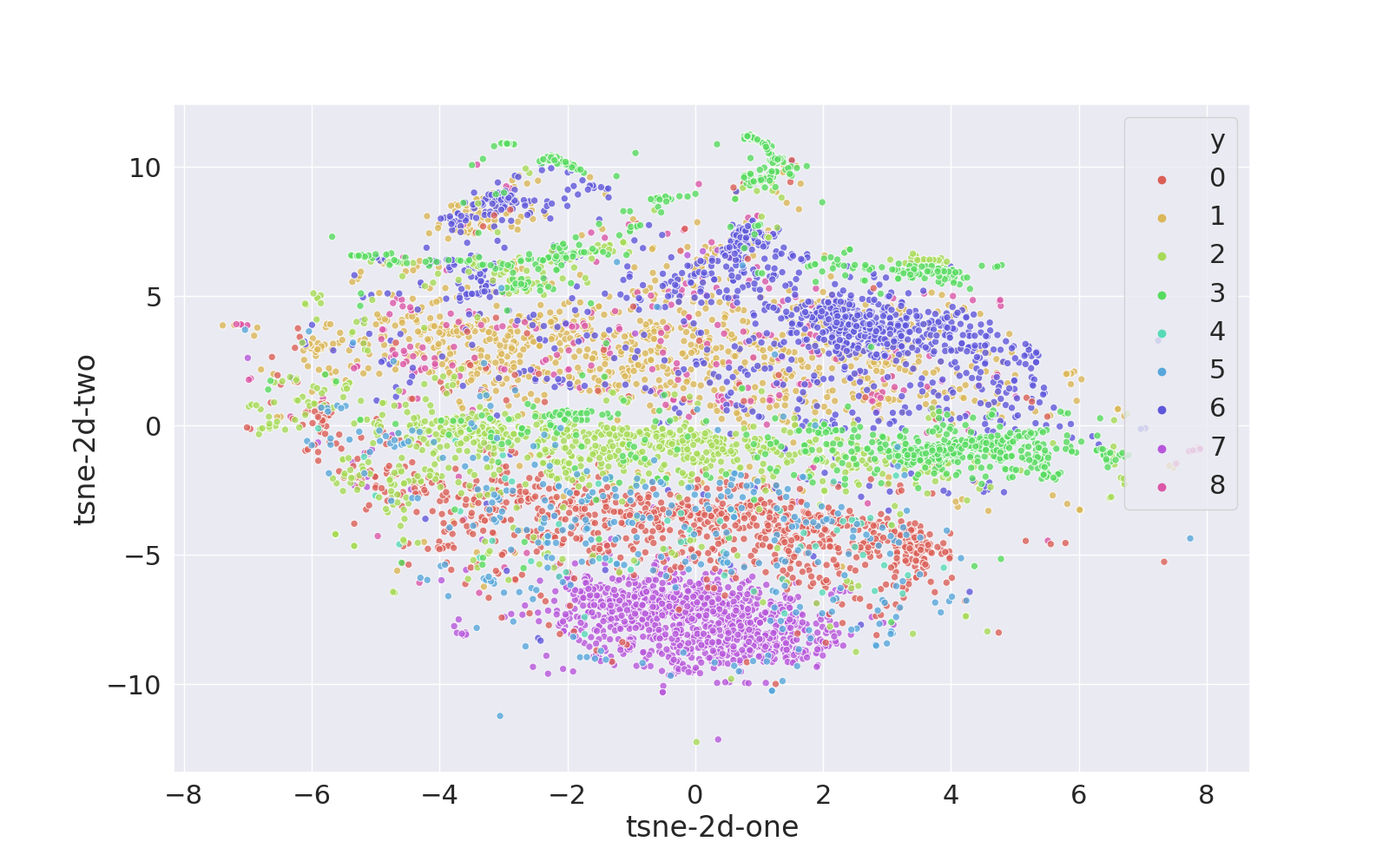}
	\caption{TSNE projection of the features extracted for CLEF-IP test dataset at the embedding layer of DGCNN.}
	\label{fig:TSNE} 
\end{figure}

\begin{figure}[htb]
	\centering 
	\includegraphics[width=0.65\linewidth]{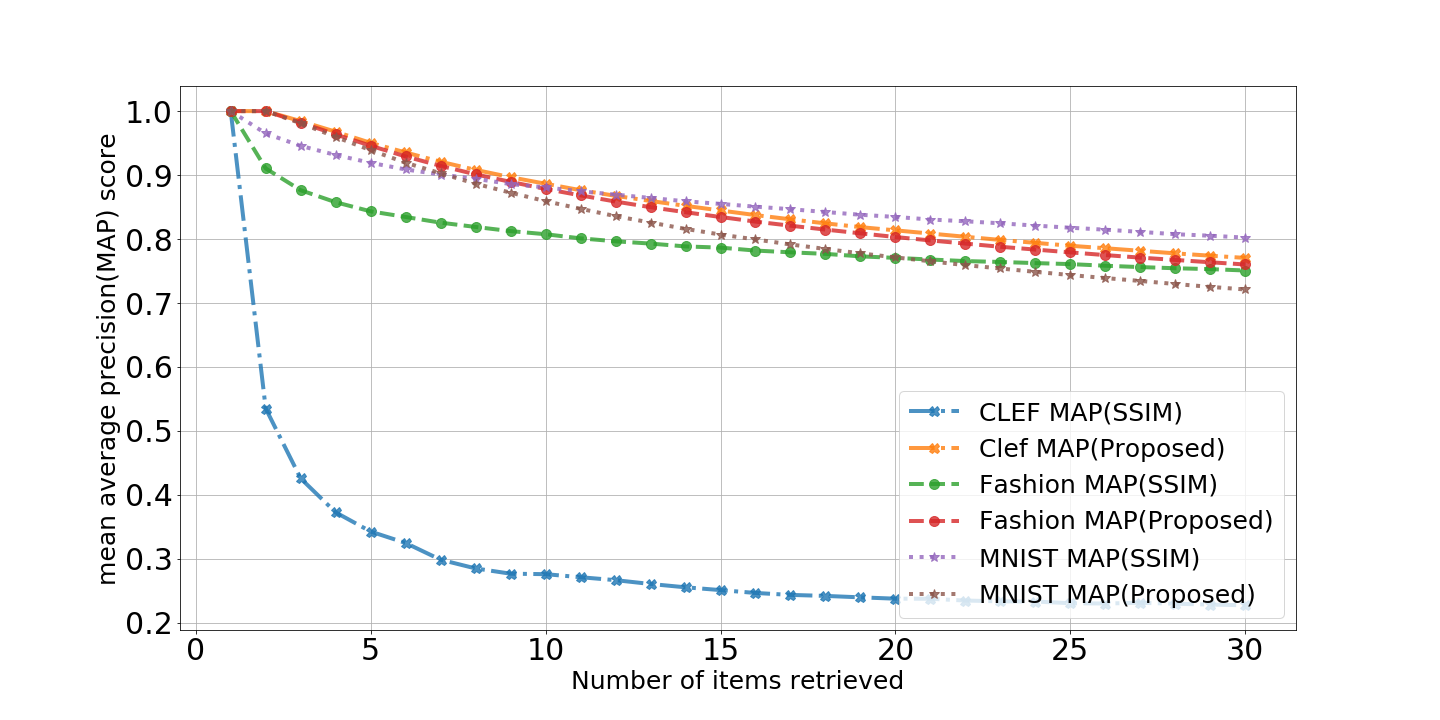}
	\caption{Mean average precision (MAP) analysis for different datasets using SSIM vs proposed approach.}
	\label{fig:retrieval} 
\end{figure}

In the case of the proposed method, we re-use the concatenated local and global features learned in the classification task after the fourth EdgeConv layer in Figure \ref{fig:architecture}. For illustrative purposes we show how the learned features naturally cluster for the CLEF-IP dataset by using the t-Distributed Stochastic Neighbor Embedding(t-SNE) \cite{maaten2008visualizing} projection. Note in Figure \ref{fig:TSNE} that features are nicely packed which seem as promising to be used to effectively retrieve similar shapes.
Standard SSIM computes the structural pair-wise similarity between all the image combinations in the test dataset partition and yields the set of $k$-images with the highest similarity measure given a query. With that, we summarize the MAP scores in both Figure \ref{fig:retrieval} and in Table \ref{tab:table_MAP} for quantitative clarity. Here, we note that the proposed method for this task outperforms SSIM in most cases, except on the MNIST dataset where there is a huge structural uniformity in similarity within class samples without significant view-point variations.
\begin{table} [htb]
	\caption{\label{tab:table_MAP} Retrieval Performances. }
	\centering
	\resizebox{0.75\columnwidth}{!}{
		\begin{tabular}{lccc}
			\hline
			\\
			\multirow{2}{*}{Dataset} &{MAP} & {MAP} & {MAP}  \\
			& @10 retrievals & @20 retrievals & @30 retrievals\\
			\hline
			\\	
			\multirow{2}{*}{MNIST} & \textbf{0.88(SSIM)} & \textbf{0.83(SSIM)} & \textbf{0.80(SSIM)} \\
			& 0.86(DGCNN) & 0.77(DGCNN) & 0.72(DGCNN) \\
			\hline    
			\\
			\multirow{2}{*}{fashion-MNIST} & 0.807(SSIM) & 0.770(SSIM) & 0.750(SSIM) \\
			& \textbf{0.878(DGCNN)} & \textbf{0.803(DGCNN)} & \textbf{0.760(DGCNN)} \\
			\hline  
			\\
			\multirow{2}{*}{CLEF-IP} & 0.276(SSIM) & 0.238(SSIM) & 0.227(SSIM)\\
			& \textbf{0.886(DGCNN)} & \textbf{0.814(DGCNN)} & \textbf{0.771(DGCNN)} \\
			%Human  &  \multirow{2}{*}{0.835} & \multirow{2}{*}{X} & X \\
			%Performance &   & &  X \\
			%Proposed &  0.849 & X & X\\
			\hline
		\end{tabular}
	}
\end{table}
% Experiments
\section{Conclusion}
\label{Sec:conclusion} 

In this work, we proposed a method connecting the successes of deep learning (DL) to alleviate some of the problems being faced in patent image classification/retrieval applications. The method combines a shape extraction pre-processing stage with a neural net operating on the spatial representation of the shape (i.e., in the non-Euclidean domain). Empirical results demonstrate the ability of the proposed method to better cope with the plausible binary image generation mechanisms which are of a highly subjective nature in patent documents. Such mechanisms included variations in drawing style, view-point, and presence of multiple components or image parts.  
%to aim solve some of the problems for shape analysis based on a graph neural net that learns features defined both on sampled shape points and interconnections between them. Our experiments on classification/retrieval tasks of multiple datasets show that the proposed method performs competitively with state of the art methods in their own domain. In future work, we envision designing a graph learning net that exploits hierarchical point-cloud multi-sparsity level features from which we expect an improved performance in more complex shape cases. In addition, we will explore in detail ways to achieve invariance to shape perspective on performance.

%%%%% No acknowledgments on a blind review paper %%%%%
\section*{Acknowledgement}
Research was supported by the Laboratory Directed Research and Development program of Los Alamos National Laboratory under project number LDRD-20200041ER.

						% Conclusion

%%%%%%%%%%%%%%%%%%%%%%%%%%%%%%%%%%%%%%%%%%%%%
%% Bibliography
%%%%%%%%%%%%%%%%%%%%%%%%%%%%%%%%%%%%%%%%%%%%%

\bibliographystyle{files/IEEEbib}
%\bibliography{files/refs}

\end{document}